\documentclass[runningheads,orivec]{llncs}

\usepackage[utf8]{inputenc}
\usepackage{geometry}
\usepackage{graphicx}
\usepackage{amsmath,amssymb,amscd,amstext}
\usepackage{xcolor}
\usepackage{pst-node}
\usepackage{tikz,pgf}
\usepackage{tikz-cd} 
\usepackage{pgfplots}
\usepackage{pgfplotstable}
\pgfplotsset{compat=newest}
\usetikzlibrary{spy,decorations.markings,decorations.text,arrows.meta,bending,3d,shapes.arrows,positioning,fit,backgrounds,fadings}

\usepackage{slashbox}
\usepackage{multirow}
\usepackage{booktabs}
\usepackage{adjustbox}
\usepackage{setspace}
\usepackage{xcolor}
\usepackage{multibib}
\usepackage{filecontents}
\usepackage{multirow}
\usepackage{booktabs}
\usepackage{mathtools}
\usepackage{commath}
\usepackage[percent]{overpic}
\usepackage[hidelinks]{hyperref}
\hypersetup{
    pdftitle={Estimate of the Neural Network Dimension using Algebraic Topology and Lie Theory},    
    pdfauthor={Luciano Melodia},     
    pdfsubject={Estimate of the Neural Network Dimension},   
    pdfcreator={Luciano Melodia},   
    pdfproducer={Luciano Melodia},  
    pdfkeywords={Embedding dimension} {Parameterization} {Persistent homology} {Neural networks} {Manifold learning}, 
    pdfnewwindow=true,
    colorlinks,
	citecolor=black,
	filecolor=black,
	linkcolor=black,
	urlcolor=blue
}
\definecolor{red1}{RGB}{227,226,223}
\definecolor{red2}{RGB}{227,175,188}
\definecolor{red3}{RGB}{238,76,124}
\definecolor{red4}{RGB}{154,23,80}
\definecolor{red5}{RGB}{93,0,30}
\makeatletter
\newcommand{\@chapapp}{\relax}
\newcommand{\xdashrightarrow}[2][]{\ext@arrow 0359\rightarrowfill@@{#1}{#2}}
\newcommand{\xdashleftarrow}[2][]{\ext@arrow 3095\leftarrowfill@@{#1}{#2}}
\newcommand{\xdashleftrightarrow}[2][]{\ext@arrow 3359\leftrightarrowfill@@{#1}{#2}}
\def\rightarrowfill@@{\arrowfill@@\relax\relbar\rightarrow}
\def\leftarrowfill@@{\arrowfill@@\leftarrow\relbar\relax}
\def\leftrightarrowfill@@{\arrowfill@@\leftarrow\relbar\rightarrow}
\def\arrowfill@@#1#2#3#4{%
  $\m@th\thickmuskip0mu\medmuskip\thickmuskip\thinmuskip\thickmuskip
   \relax#4#1
   \xleaders\hbox{$#4#2$}\hfill
   #3$%
}

\makeatother
\begin{document}
\title{Estimate of the Neural Network Dimension using Algebraic Topology and Lie Theory}
%
%
\titlerunning{Estimate of the Neural Network Dimension}

\author{Luciano Melodia\orcidID{0000-0002-7584-7287} \\
\and Richard Lenz\orcidID{0000-0003-1551-4824}
}%
\authorrunning{L. Melodia et al.}
%
\institute{Chair of Computer Science 6\\
Friedrich-Alexander University Erlangen-Nürnberg\\
91058 Erlangen, Deutschland \\
\email{\{luciano.melodia,richard.lenz\}@fau.de}}
\maketitle              
\begin{abstract}
In this paper we present an approach to determine the smallest possible number of neurons in a layer of a neural network in such a way that the topology of the input space can be learned sufficiently well. We introduce a general procedure based on persistent homology to investigate topological invariants of the manifold on which we suspect the data set. We specify the required dimensions precisely, assuming that there is a smooth manifold on or near which the data are located. Furthermore, we require that this space is connected and has a commutative group structure in the mathematical sense. These assumptions allow us to derive a decomposition of the underlying space whose topology is well known. We use the representatives of the $k$-dimensional homology groups from the persistence landscape to determine an integer dimension for this decomposition. This number is the dimension of the embedding that is capable of capturing the topology of the data manifold. We derive the theory and validate it experimentally on toy data sets. 

\end{abstract}

\section{Motivation}
Since the development of deep neural networks, their parameterization, in particular the smallest possible number of neurons in a layer, has been studied. This number is of importance for auto-encoding tasks that require extrapolation or interpolation of data, such as blind source separation or super-resolution. If this number is overestimated, unnecessary resources are consumed during training. If the number is too small, no sufficiently good estimate can be given. To solve this problem, we make the following contribution in this paper:
\begin{itemize}
    \item We study topological invariants using statistical summaries of persistent homology on a filtered simplicial complex on the data.
    \item Using the theory of Lie groups, we specify the smallest possible embedding dimension so that the neural network is able to estimate a projection onto a space with the determined invariants.
\end{itemize}

\section{Related Work}
We present relevant work on neural networks from a differential geometric perspective. Further, we summarize earlier attempts to determine depth and width of a neural network, depending on the data to be processed.

\subsection{Differential Geometry in Neural Networks}
The manifold of a dense neural network is in most cases Euclidean. Nevertheless, it does often approximate a manifold with different structure. Recently there have been results in the direction of other manifolds on which neural networks can operate and that fit more the nature of data, such as spherical neural networks \cite{CohenGKW18}. They operate on a model of a manifold with commutative group structure.

Manifolds induce a notion of distance, with symmetric properties, also referred to as metric. A metric describes the geometric properties of a manifold. The change of coordinate systems and the learned metric tensor of a smooth manifold during back propagation was formalized \cite{HauserR17}, which shows, that we often operate on a model of a Lie group. Further, it has been shown that dense neural networks can't approximate all functions arbitrarily precisely \cite{Johnson19}. Thereupon, neural networks with residual connections were investigated. Residual networks add the output of one layer to a deeper one, bypassing some of the layers in between. Indeed, they are universal function approximators \cite{LinJ18}. Inspired by finite difference methods, such a residual layer was defined as a forward or backward difference operator on a partition of layers. Its state space dimension with residual connections is homeomorphic to $\mathbb{R}^{k \cdot n}$, where $n = \text{dim} \; M$ and $k$ is the number of times the difference operator was used \cite{HauserGJR19}. Such a layer is able to embed into $(k\cdot n)$-dimensional Euclidean space. The smooth manifold perspective inspired us to investigate its invariants.

\subsection{Embedding Dimension of Neural Networks}
Cybenko showed that dense neural networks with one hidden layer can approximate any continuous function with a bounded domain with arbitrarily small error \cite{Cybenko92}. Raghu et al. \cite{RaghuPKGS17} quantified the upper bound for \texttt{ReLU} and \texttt{hard tanh} networks considering the length of activation patterns -- a string of the form $\{0,1\}^k$ for \texttt{ReLU}-networks and $\{-1,0,1\}^k$ for \texttt{hard tanh}-networks -- with $k$ being the number of neurons. They gave a tight upper bound for both in the context of dense networks. The activation pattern for \texttt{ReLU}-networks is bounded by $\mathcal{O}(k^{mn})$ and the one for \texttt{hard tanh} is bounded by $\mathcal{O}((2k)^{mn})$, with $m = \dim \mathbb{R}^m$ being the dimension of input space, $n$ being the number of hidden layers and $k$ being their width, respectively. Bartlett et al. show an almost tight bound for VC dimensions. A neural network with $W$ weights and $L$ layers is bounded by $O(WL \log W)$ VC dimensions but has $\Omega(WL \log W/L)$ VC dimensions \cite{BartlettHLM19}. Its depth was determined as a parameter depending on the moduli of continuity of the function to be approximated \cite{LinJ18}.

To these fundamental results we contribute a topological approach to the parameterization of the minimal amount of neurons within a layer, such that the topology of the data manifold can be represented. Similar to Futagami et al.\cite{FutagamiYS19}, we investigate the manifold on which we suspect the data. Our approach is applicable to topological spaces in general and is based on persistent homology. We look for the dimension of a space with the same topological properties that the data indicates and which can be approximated. We assume a simple decomposition into a product space of real planes and $1$-spheres to get the same homology groups as the persistent ones on a filtration of data. The dimension of the decomposition is a lower bound.

\section{The Manifold Assumptions}
The assumption that a set of points lies on a manifold is more accurate than the treatment in Euclidean space. A topological manifold $M$ is a Hausdorff space, which means that any two points $x,y \in M$ always have open neighborhoods $U_x,U_y \subset M$, so that their intersection is empty. This behaviour may give an intuitive feeling for the position of points in space. The topology $(M, \nu)$ is a set system that describes the structure of a geometric object. Here $M$ itself and the empty set $\emptyset$ must be contained in $\nu$ and any union of open sets from $\nu$ and any intersection of $\nu$ must be contained in $\nu$. If this set system has a subset which generates the topology $\nu$ by any unions, then this subset is a basis of the topology. We demand of a manifold that the basis is at most countable. In addition, $M$ is locally Euclidean, i.e. each point has a neighborhood which can be mapped homeomorphically to a subset of $\mathbb{R}^n$. The natural integer $n$ is the dimension of the manifold $M$. 

\subsection{Smooth Manifolds}
A smooth structure is a stronger condition than a topological manifold which can be described by a family of continuous functions. The assumption to operate on a smooth manifold when processing data is justified by the theorem of Stone-Weierstrass which proves that every continuous function can be approximated arbitrarily exactly by a specific smooth one, namely a polynomial \cite{stone1948generalized}. Smooth manifolds are described by a family of local coordinate maps $\varphi: U \rightarrow \varphi(U) \subseteq \mathbb{R}^n$ which are homeomorphic to a subset of $\mathbb{R}^{n}$ and cover an open neighborhood $(U_i,\varphi_i)$ in $M$. A family of charts $\mathcal{A} = (U_i,\varphi_i)_{i \in I}$ is an atlas on $M$ \cite{lee2013smooth}. $x_1, \cdots, x_n: U \rightarrow \mathbb{R}$ are local coordinates $\varphi(p) = (x_1(p), \cdots, x_n(p))$. If the atlas is maximal in terms of inclusion, then it is a differentiable structure. Each atlas for $M$ is included in a maximal atlas. Due to the differentiable structure, the maps in $\mathcal{A}$ are also compatible with all maps of the maximal atlas. Thus, it is sufficient to use a non maximal atlas.

As long as the activation functions of a neural network are of the same differentiable structure, they also are compatible with each other, and the manifold does not change as the data propagates through the layers. Only the coordinate system changes.

\subsection{Lie Groups}
The study of group theory deals with symmetries which can be expressed algebraically. A pair $(M,\circ)$, consisting of a map $\circ: M\times M \rightarrow M$ is called group if the map is associative, i.e. $x \circ (y \circ z) = (x \circ y) \circ z$ and has a neutral element so that $x \circ e_{M} = x$. We also require an inverse element to each element of the group, i.e. $x \circ x^{-1} = e_{M}$. A group is abelian if all elements commute under the group operation. A Lie group is a smooth manifold which is equipped with a group structure such that the maps $\circ: M \times  M \rightarrow  M, (x,y) \mapsto xy, $ and $\imath: M \rightarrow  M, x \mapsto x^{-1}$ are smooth. We call a space connected if it can't be divided into disjoint open neighbourhoods. We apply a theorem from Lie theory, which states that each connected abelian Lie group $(M,\circ)$ of dimension $\dim M = n$ is isomorphic to a product space $\mathbb{T}^q \times \mathbb{R}^p $ with $p+q =n$, for a proof we refer to \cite[p.~116]{onishchik1993lie}. The $q$-torus $\mathbb{T}^q$ is a surface of revolution. It moves a curve around an axis of rotation. These curves are given by $1$-spheres, such that the $q$-torus is a product space $\mathbb{T}^q = S^1_1 \times \cdots \times S^1_q$. The initial decomposition of connected commutative Lie groups can be further simplified into $M \cong \mathbb{R}^p \times S^1_1 \times \cdots \times S^1_q$. Recall, that the $(n-1)$-sphere is given by $S^{n-1} := \{x \in \mathbb{R}^n \; \vert \; \norm{x}_2 = 1\}$. Thus, $S^1$ can be embedded into $\mathbb{R}^2$.

Next, we derive how many dimensions are at least needed for a suitable embedding. For each $1$-sphere we count two dimensions and for each real line accordingly one. Finally, we have to estimate from data how the decomposition may look like to yield the topology of the data manifold.

\section{Persistent Homology}
Algebraic topology provides a computable tool to study not the topology of a set of points themselves, but an abelian group attached to them. The core interest in this discipline lies in homotopy equivalences, an equivalence class to which objects belong that are continuously deformable into one another. This is much broader than homeomorphism. For two topological spaces $X$ and $Y$ we seek for a function $h: X \times I \rightarrow Y$, which gives the identity at time $h_0(X) = X$ and for $h_1(X) = Y$ a mapping into another topological space. If this mapping is continuous with respect to its arguments, it is called homotopy. Consider two functions $f,g: I \rightarrow X$ so that $f(1) = g(0)$. Then there is a composition of product paths $f \cdot g$, which pass first through $f$ and then through $g$ and which is defined as $f \cdot g (s) = f(2s)$ for $0 \leq s \leq 1/2$ and $g(2s - 1)$, for $1/2 \leq s \leq 1$. We first run $f$ at double speed up to $1/2$ and from $1/2$ to $1$ we run the function $g$ at double speed. In addition, suppose that a family of functions $f: I \rightarrow X$ is given which have the same start and end point $f(0) = f(1) = x_0 \in X$. They intuitively form a loop. The set of all homotopy classes $[f]$ of loops $f: I \rightarrow X$ at base point $x_0$ is noted as $\pi_1(X,x_0)$ and is called first homotopy group or fundamental group. The group operation is the product of equivalence classes $[f][g] = [f \cdot g]$. If we do not look at the interval $I$, but at mappings considering the unit cube $I^n$, we obtain the $n$-th homotopy group $\pi_n(X,x_0)$ by analogy. With the help of homotopy groups we study connected components of a topological space for the $0$-th group, the loops for the $1$-st group, the cavities for the $2$-nd and so forth. Since homotopy groups are difficult to compute, we resort to an algebraic variant, the homology groups.

\subsection{Simplices}
In data analysis, we do not study the topological spaces themselves, but points that we assume are located on or near this space. In machine learning we mostly deal with the investigation of closed surfaces, which can always be triangulated, i.e. completely covered with simplices. For a proof with excellent illustrations we refer to \cite[p.~102]{hatcher2002algebraic}. The concept of a triangle is too specific for our purpose, thus we'll define the $n$-simplices as a generalization. They are the smallest convex set in Euclidean space with $(n+1)$-points $v_0,v_1,\cdots,v_n$, having no solutions for any system of linear equations in $n$-variables. Thus, we say they lie in general position with respect to $\mathbb{R}^n$, because they do not lie on any hyperplane with dimension less than $n$ \cite[p.~103]{hatcher2002algebraic}. We define the $n$-simplex as follows:
\begin{equation}
    \label{simplex}
    \sigma = [v_0,...,v_n] = \left\{ v_i = \sum_{j=0}^{n} \lambda_j v_j \; \bigg\vert \; \sum_{j=0}^{n} \lambda_j = 1 \; \text{and} \; \lambda_j \geq 0 \; \text{for all} \; j \right\}.
\end{equation}
Removing a vertex from $\sigma$ results in a $(n-1)$-simplex called face of $\sigma$. A $0$-simplex is a point, a $1$-simplex is a path between two $0$-simplices, a $2$-simplex is a surface enclosed by three $1$-simplices and so on. Therefore, they follow the intuition and generalize triangles to polyhedra, including points by definition.
\begin{figure}[t]
    \begin{overpic}[width=\textwidth,tics=10]{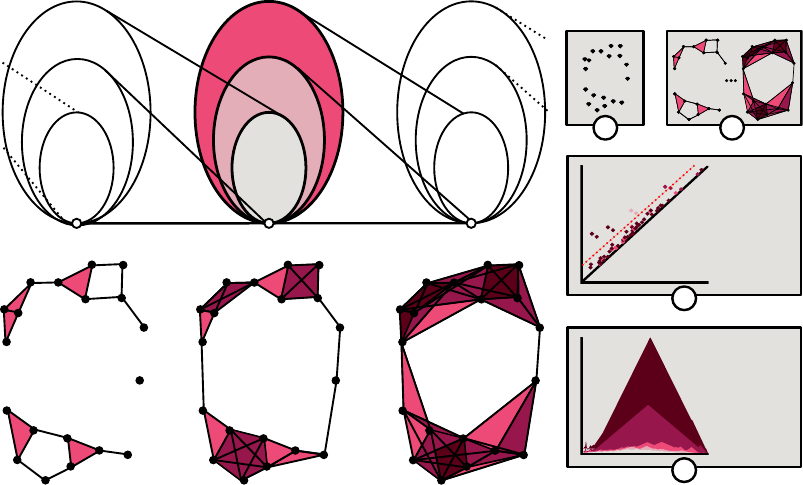}
        \put (0.9,58) {\textbf{a})}
        \put (0.9,28) {\textbf{b})}
        \put (70,58) {\textbf{c})}
        \put (8.8,29.6) {$0$}
        \put (32.8,29.6) {$0$}
        \put (58,29.6) {$0$}
        \put (74.8,43.6) {\small{1}}
        \put (90.5,43.6) {\small{2}}
        \put (84.6,22.4) {\small{3}}
        \put (84.6,1) {\small{4}}
        \put (7,13) {$K_1$}
        \put (31,13) {$K_2$}
        \put (57,13) {$K_3$}
        \put (21,13) {$\subseteq$}
        \put (45,13) {$\subseteq$}
        \put (32,39) {$B_k$}
        \put (32,49) {$Z_k$}
        \put (32,55) {$C_k$}
        \put (56,39) {$B_{k-1}$}
        \put (56,49) {$Z_{k-1}$}
        \put (56,55) {$C_{k-1}$}
        \put (7,39) {$B_{k+1}$}
        \put (7,49) {$Z_{k+1}$}
        \put (7,55) {$C_{k+1}$}
        \put (19.5,56) {$\partial_{k+1}$}
        \put (44.5,56) {$\partial_{k}$}
        \put (92,37) {{\color{red5} $\bullet$} $H_0$}
        \put (92,34) {{\color{red4} $\bullet$} $H_1$}
        \put (92,31) {{\color{red3} $\bullet$} $H_2$}
        \put (92,28) {{\color{red2} $\bullet$} $H_3$}
        \put (92,25) {$\cdots$}
        \put (92,15.5) {{\color{red5} $\bullet$} $H_0$}
        \put (92,12.5) {{\color{red4} $\bullet$} $H_1$}
        \put (92,9.5) {{\color{red3} $\bullet$} $H_2$}
        \put (92,6.5) {{\color{red2} $\bullet$} $H_3$}
        \put (92,3.5) {$\cdots$}
    \end{overpic}\\[0.1cm]
    \caption{\textbf{a}) Illustration of the chain complex following Zomorodian et al. \cite{ZomorodianC05}. \textbf{b}) Sublevel sets of a simplicial complex connected by inclusion. \textbf{c}) Pipeline for persistent homology: \textbf{c})1. Loading point sets. \textbf{c})2. Computation of a filtration. \textbf{c})3. Computation of persistent homology with confidence band. \textbf{c})4. Computation of persistence landscapes.}
    \label{figure1}
\end{figure}
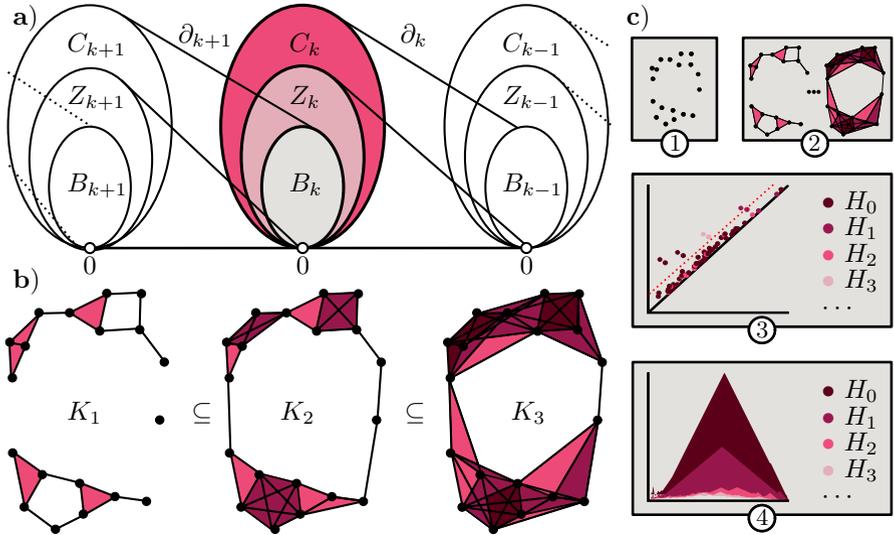

\subsection{Simplicial Complexes}
A finite simplicial complex $K$ is a set of simplices, such that any face of a simplex of $K$ is a simplex of $K$ and the intersection of any two simplices of $K$ is either empty or a common face of both \cite[p.~11]{boissonnat2018geometric}. This defines the simplicial complex as a topological space. Illustrations for simplicial complexes can be found in Fig. \ref{figure2} b). Their properties ensure that simplices are added to the complex in a way, such that they are `glued' edge to edge, point to point, and surface to surface. Thus, every $n$-simplex has $n+1$ distinct vertices and no other $n$-simplex has the same set of vertices. This is a unique combinatorial description of vertices together with a collection of sets $\{\sigma_1, \sigma_2, \cdots, \sigma_m\}$ of $k$-simplices, which are $k+1$ element subsets of $K$ \cite[p.~107]{hatcher2002algebraic}. This set system can be realized geometrically with an injective map into Euclidean space or any other metric space. We would like to explain in brief, using a plain example, how we can create a simplicial complex from a set of points. For this purpose, we will not use the Delaunay complex, which will be utilised in the subsequent experiments, but will base our remarks on the Vietoris-Rips complex, which fulfils the explanatory purposes.

Let $X = \{x_0, x_1, \cdots, x_n\}$ be a set of points in general position sampled from $\mathbb{R}^{n+1}$. Then the Vietoris-Rips complex contains all points $\{x_0,x_1, \cdots, x_k\}$ $\in$ $\text{Rips}^\epsilon(X)$ for which $\vert\vert x_i - x_j \vert\vert \leq \epsilon$. The simplicial complexes for increasing $\epsilon$ are connected by inclusion and form a filtration, see Sect. \ref{homologicalpersistence}.

\subsection{Associated Abelian Groups}
Algebraically, the simplices can be realized as a system of linear combinations as $\sum_i \lambda_i \sigma_{i}$ following Eq. \ref{simplex}. Together with addition as an operation, they form a group structure called $k$-th chain group $(C_k, +)$ for the corresponding dimension of the $k$-simplices used. The group is commutative or abelian. The used coefficients $\lambda_i$ induce an orientation, intuitively a direction of the edges, by negative and positive signs, such that $\sigma = -\tau$, iff $\sigma \sim \tau$ and $\sigma$ and $\tau$ have different orientations. The objects of the chain group are called $k$-chains, for $\sigma \in C_k$. The groups are connected by a homomorphism, a mapping which respects the algebraic structure, the boundary operator:
\begin{align}
\partial_k: C_k &\rightarrow C_{k-1}, \\
\partial_k \sigma &\mapsto \sum_i (-1)^i [v_0,v_1,\cdots,\hat{v}_i, \cdots, v_n].
\end{align}
The $i$-th face is omitted, alternatingly. Using the boundary operator we yield the following sequence of abelian groups, which is a chain complex (cf. Fig. \ref{figure1} a)):
\begin{equation}
    0 \xrightarrow[]{\partial_{k+1}} C_k \xrightarrow[]{\partial_{k}} C_{k-1} \xrightarrow[]{\partial_{k-1}} \cdots \xrightarrow[]{\partial_{2}} C_1 \xrightarrow[]{\partial_{1}} C_0 \xrightarrow[]{\partial_{0}} 0.
\end{equation}
For the calculation of homology groups we have to choose some algebraic structure from which to obtain the coefficients $\lambda_i$. Since, according to the Universal Coefficient Theorem, all homology groups are completely determined by homology groups with integral coefficients, the best choice would be the ring of integers $(\mathbb{Z},+,\cdot)$, see the proof in \cite{gruenberg1968universal}. A field, however, has more favourable arithmetic properties, so by convention $\mathbb{Z}_p := \mathbb{Z}/p\mathbb{Z}$ is chosen for a prime $p$ generating a maximal ideal. For our experiments we have therefore chosen $\mathbb{Z}_2$.

The $k$-th chain groups contain two different abelian subgroups that behave normal to it. First, the so-called cycle groups $(Z_k, +)$, which are defined as
\begin{align}
Z_k := \ker \partial_k = \{\sigma \in C_k \; \vert \; \partial_k \sigma = \emptyset \}.
\end{align}
This follows the intuition of all elements that form a loop, i.e. have the same start and end point. Some of these loops have the peculiarity of being a boundary of a subcomplex. These elements form the boundary group $(B_k, +)$, defined by
\begin{align}
B_k := \text{im} \; \partial_{k+1} = \{\sigma \in C_k \; \vert \; \exists \tau \in C_{k+1}: \sigma = \partial_{k+1}\tau\}.
\end{align}
The connection of these groups is illustrated in Fig. \ref{figure1} a). Now we define the homology groups as quotients of groups. Homology -- analogous to homotopy theory -- gives information about connected components, loops and higher dimensional holes in the simplicial complex:
\begin{equation}
    H_k(K) := \frac{\ker \partial_k C_k(K)}{\text{im } \partial_{k+1} C_{k+1}(K)} = \frac{Z_k(K)}{B_k(K)}.
\end{equation}

\subsection{Homological Persistence}
\label{homologicalpersistence}
Examining the homology groups of a set of points gives little information about the structure of a data set. Instead, we are interested in a parameterization of the simplicial complex as geometric realization in which the homology groups appear and disappear again. For this purpose we consider all possible subcomplexes that form a filtration, a nested sequence of subcomplexes, over the point set $X$. We denote $K^{\epsilon} := K^{\epsilon}(X)$. Depending on how we vary the parameter $\epsilon_i$ of the chosen simplicial complex, the following sequence is generated, connected by inclusion:
\begin{align}
    \emptyset &= K^{\epsilon_0} \subseteq K^{\epsilon_1} \subseteq \cdots \subseteq K^{\epsilon_{n+1}}= K,\\
    K^{\epsilon_{i+1}} &= K^{\epsilon_i} \cup \sigma^{\epsilon_{i+1}}, \quad \text{for} \; i \in \{0,1, \cdots, n-1\}.
\end{align}
An example of a filtration is given in Fig. \ref{figure1} b). The filtration has in our case a discrete realization with a heuristically fixed $\epsilon = \epsilon_{i+1}-\epsilon_{i} = \min(||x-y||_2)$ for all $x,y \in X$.

Through filtration we are able to investigate the homology groups during each step of the parameterization. We record when elements from a homology group appear and when they disappear again. Intuitively speaking, we can see when $k$-dimensional holes appear and disappear in the filtration. We call this process birth and death of topological features. Recording the Betti numbers of the $k$-th homology group along the filtration, we obtain the $k$-dimensional persistence diagram (see Fig. \ref{figure1} c)3). The Betti numbers $\beta_k$ are defined by $\text{rank } H_k$. We write $H^{\epsilon_i}_k$ as $k$-th homology group on the simplicial complex $K$ with parameterization $\epsilon_i$. Then $H^{\epsilon_{i}}_{k} \rightarrow H^{\epsilon_{i+1}}_k$ induces a sequence of homomorphisms on the filtration, for a proof we refer to \cite{edelsbrunner2008persistent}:
\begin{align}
0 = H^{\epsilon_{0}}_k \rightarrow H^{\epsilon_{1}}_k \rightarrow \cdots \rightarrow H^{\epsilon_{n}}_k \rightarrow H^{\epsilon_{n+1}}_k = 0.
\end{align}
The image of each homomorphism consists of all $k$-dimensional homology classes which are born in $K^{\epsilon_i}$ or appear before and die after spanning $K^{\epsilon_{i+1}}$. Tracking the Betti numbers on the filtration results into a multiplicity
\begin{equation}
\mu^{\epsilon_{i},\epsilon_{j}}_k = (\beta_k^{\epsilon_{i},\epsilon_{j-1}} - \beta_k^{\epsilon_i,\epsilon_{j}})-(\beta_k^{\epsilon_{i-1},\epsilon_{j-1}}-\beta_k^{\epsilon_{i-1},\epsilon_{j}}),
\end{equation}
for the $k$-th homology group and index pairs $(\epsilon_i,\epsilon_{j+1}) \in \overline{\mathbb{R}^2_+} := \mathbb{R}^2_+ \cup \{\infty,\infty\}$ with indices $i \leq j$. The Euclidean space is extended, as the very first connected component on the filtration remains connected. Thus, we assign to it infinite persistence, corresponding to the second coordinate $\epsilon_{j+1}$. The first term counts elements born in $K^{\epsilon_{j-1}}$ and which vanish entering $K^{\epsilon_{j}}$, while the second term counts the representatives of homology classes before $K^{\epsilon_{j}}$ and which vanish at $K^{\epsilon_{j}}$. The $k$-th persistence diagram is then defined as
\begin{equation}
\text{Ph}_{k}(X) := \left\{(\epsilon_i, \epsilon_{j+1}) \in \overline{\mathbb{R}^2_+} \; \bigg\vert \; \mu^{\epsilon_{i},\epsilon_{j+1}}_k = 1, \; \forall \; i,j \in \{0,1,\cdots,n-1\} \right\}.
\end{equation}

\subsection{Persistence Landscapes}
Persistence landscapes give a statistical summary of the topology of a set of points embedded in a given topological manifold \cite{bubenik2015statistical}. Looking at the points $(\epsilon_i,\epsilon_{j+1}) \in \overline{\mathbb{R}^2_+}$ on the $k$-th persistence diagram $\text{Ph}_{k}(X)$, one associates a piecewise linear function $\lambda^{\epsilon_i}_{\epsilon_{j+1}}: \mathbb{R} \rightarrow [0,\infty)$ with those points:
\begin{align}
    &\text{If } x \not\in (\epsilon_i,\epsilon_{j+1}), \; \lambda^{\epsilon_i}_{\epsilon_{j+1}}(x) = 0,\\
    &\text{if } x \in \left(\epsilon_{i},(\epsilon_i + \epsilon_{j+1})/2 \right], \; \lambda^{\epsilon_i}_{\epsilon_{j+1}}(x) = x - \epsilon_{i} \; \text{and}\\
    &\text{if } x \in \left((\epsilon_i + \epsilon_{j+1})/2,\epsilon_{j+1} \right), \; \lambda^{\epsilon_i}_{\epsilon_{j+1}}(x) = \epsilon_{j+1} - x.
\end{align}
The summaries for the general persistence diagram are the disjoint union of the $k$-th persistence diagrams $\text{Ph}(X) := \coprod_{i=0}^{k} \text{Ph}_i(X)$. A persistence landscape $\text{Pl}(X)$, contains the birth-death pairs $(\epsilon_i,\epsilon_{j+1})$, for an $i,j \in \{1, \cdots, n\}$ and is the function sequence $\Lambda_k : \mathbb{R} \rightarrow [0,\infty)$ for a $k \in \mathbb{N}$, where $\Lambda_k(x)$ denotes the $k$-th greatest value of $\lambda^{\epsilon_i}_{\epsilon_{j+1}}(x)$ (see Fig. \ref{figure1} c)4). Thus, $\Lambda_k(x) = 0$ if $k > n$. $\text{Pl}(X)$ lies in a completely normed vector space, suited for statistical computations.

\section{Neural Networks}
\label{sec:neuralnets}
Neural networks are a composition of affine transformations with a non-linear activation function. This transformation obtains collinearity. It also preserves parallelism and partial relationships. The projection onto the $(l+1)$-th layer of such a network can be written as 
\begin{align}
\textbf{x}^{(l+1)} = f(\mathbf{W}^{(l+1)} \cdot \textbf{x}^{(l)}+\textbf{b}^{(l+1)}).
\end{align}
The composition of multiple such maps is a deep neural network. The linear map $\textbf{x} \mapsto \mathbf{W} \textbf{x}$ and the statistical distortion term $\textbf{x} \mapsto \textbf{x}+\textbf{b}$ are determined by stochastic gradient descent. Note, that the linear transformation $\textbf{x} \mapsto \mathbf{W} \cdot \textbf{x}$ can be interpreted as the product of a matrix $\mathbf{W}_{ij} = \mathbf{\delta}_{ij}\mathbf{W}_i$ with the input vector $\textbf{x}$ using the Kronecker-$\delta$, while $(\cdot)$ denotes element-wise multiplication.

Commonly used are functions from the exponential family. As a result a neural network is a map $\varphi^{(l)}:\textbf{x}^{(l)}(M)$ $\rightarrow$ $(\varphi^{(l)} \circ \textbf{x}^{(l)})(M)$. Learning by back propagation can in this way be considered as a change of coordinate charts of a smooth manifold. For a detailed formulation of back propagation as a shift on the tangent bundle of this manifold we refer to Hauser et al. \cite{HauserR17}. In practice, neural networks are used with a different number of neurons per layer. However, a layer can represent the manifold of the previous one. Manifolds can always be immersed and submersed as long as the rank of the Jacobian of the respective map does not change. Thus, the dimension of the data manifold is equal to the width of the smallest possible layer of a neural network that is able to represent it \cite{HauserR17}. In other words, the layer with minimal width which does not loose topological structure.

\section{Counting Betti Numbers}
According to our main assumption, we seek a decomposition of the data manifold into a product of the real plane and tori. We refer to this Lie group as $G \cong \mathbb{T}^q \times \mathbb{R}^p$. Thus, there is also an isomorphism of the homology groups of $G$ with those of $\mathbb{T}^q \times \mathbb{R}^p$, so that
\begin{align}
H_k(G) \cong H_k(S^{1}_{1}\times\cdots\times S^{1}_{q} \times \mathbb{R}^p),
\end{align}
with $q+p = \dim G$. We demand from a neural network that these homology groups can all be approximated, i.e. that all topological structure of the input space can be represented sufficiently well, by means of its topological invariants.

Using Künneth's theorem, we know that the $k$-th singular homology group of a topological product space is isomorphic to the direct sum of the tensor product of $k$-th homology groups from its factors, for a proof see \cite[p.~268]{hatcher2002algebraic}:
\begin{align}
    H_k(X \times Y) \cong \bigoplus_{i+j=k}H_i(X) \otimes H_j(Y).
\end{align}
This applies to all fields, since modules over a field are always free. If we apply the theorem to the decomposition, we get for the desired space
\begin{align}
    \label{homologykunneth}
    &H_k(S^{1}_{1}\times\cdots\times S^{1}_{q}\times\mathbb{R}^p) \cong \\
    \bigoplus_{i_{1} + \cdots + i_{r}=k} &H_{i_{1}}(S^{1}_{1}) \otimes \cdots \otimes H_{i_{r-1}}(S^{1}_{q}) \otimes H_{i_{r}}(\mathbb{R}^{p}).
\end{align}

One might wonder to what extent homology groups of a simplicial complex can be used to estimate homology groups of smooth manifolds. Through construction of continuous maps from a simplex into a topological space, so-called singular simplices, the singular homology groups $H_{k}(X)$ are obtained. We refer to Hatcher \cite[p.~102]{hatcher2002algebraic} for a proof. This allows to assign abelian groups to any trianguliable topological space. If one constructs smooth mappings instead of continuous ones, one gets smooth singular homology groups $H^{\infty}_{k}(X)$, consider \cite[pp.~473]{lee2013smooth} for a proof. Finally a chain complex can be formed on a smooth manifold over the $p$-forms on $X$ and from this the so-called de Rham cohomology can be defined $H_{\text{dR}}^{k}(X)$, for a proof we refer to \cite[pp.~440]{lee2013smooth}. In summary, we have homology theories for simplicial complexes, topological spaces and for smooth manifolds. Simplicial homology is determined by a simplicial complex, which is merely a rough approximation of the triangulation of the underlying topological space. However, we can draw conclusions about a possible smooth manifold, since the discussed homology groups all have isomorphisms to each other known as the de Rham Theorem, see \cite[pp.~106, pp.~467]{hatcher2002algebraic,lee2013smooth}. By the isomorphism of simplicial, singular, smooth singular and de Rham cohomology, our approach is legitimized.

We count the representatives of homology classes in the persistence landscapes to derive the dimension of the sought manifold. We get for $S^1$:
\begin{align}
    H_0(S^1) &\cong H_1(S^1) \cong \mathbb{Z},\\
    H_i(S^1) &\cong 0, \; \text{for all } i \geq 2.
\end{align}
Note, in Eq. \ref{homologykunneth} terms remain only for indices $i_{j} \in \{0,1\}$. Thus, we get
\begin{align}
    H_0(\mathbb{R}^p) &\cong\mathbb{Z},\\
    \label{determineq}
    H_k(\mathbb{T}^{q}) &\cong H_k(S^{1}_{1} \times \cdots \times S^{1}_{q}) \cong \mathbb{Z}^{{q\choose{k}}},
\end{align}
where $p$ indicates the number of connected components. The persistence landscape functions $\lambda^{\epsilon_i}_{\epsilon_{j+1}}(x)$ are in general not differentiable for an $x$ yielding a maximum value. For a smooth approximation $\tilde{\lambda}$, of each specific $\lambda$, the number of local maxima for a homology group in the persistence landscape is the cardinality of the set of points $(\epsilon_i,\epsilon_{j+1})$ with derivatives $d\tilde{\lambda}^{\epsilon_i}_{\epsilon_{j+1}}(x) = 0 \; \text{and} \; d^2 \tilde{\lambda}^{\epsilon_i}_{\epsilon_{j+1}}(x) < 0$, for an example of these maximum values within a persistence landscape (see Fig. \ref{figure2} b), c)).

Thus, it is the solution for the binomial coefficient:
\begin{align}
     {q \choose k} = \frac{q}{1} \cdot \frac{q-1}{2} \cdots \frac{q-(k-1)}{k} = \prod_{i=1}^{k} \frac{q+1-i}{i}.
\end{align} 
We count in $\text{Pl}(X)$ the elements of the $0$-th homology group. The amount of elements from higher homology groups correspond to $q$ (Eq. \ref{determineq}) and are computed using brute force for (approximate) integer solutions (see Tab. \ref{figure3}).

\section{Experimental Setting}
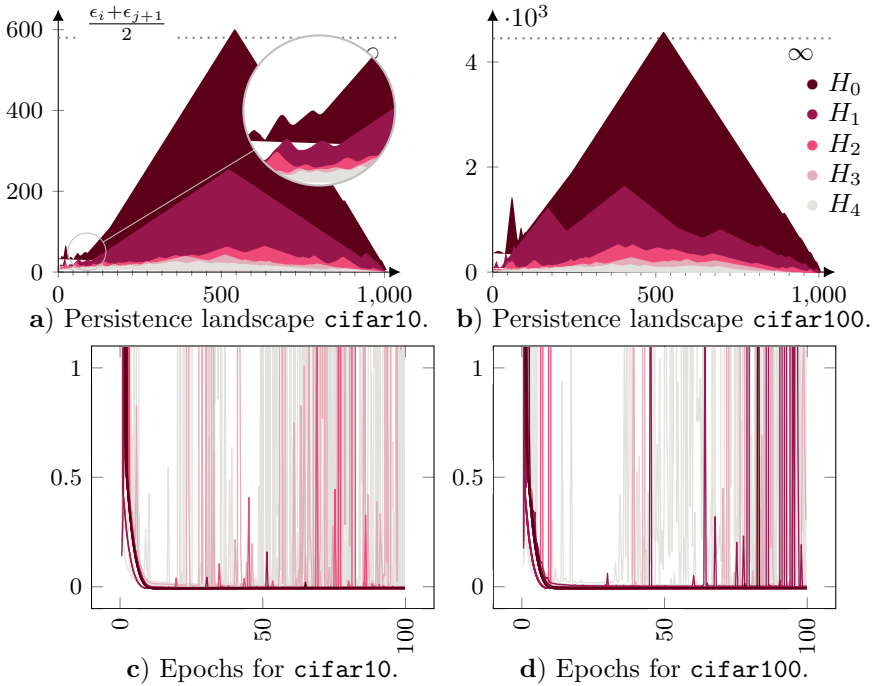
\begin{figure}[t!]
\centering
\begin{tikzpicture}
\begin{axis}[
	height=2in,
	width= 2.4in,
	axis x line = bottom,
	axis y line = left,
	minor x tick num=10,
	ylabel={$\frac{\epsilon_i + \epsilon_{j+1}}{2}$},
	xlabel={\textbf{a}) Persistence landscape \texttt{cifar10}.},
	xmin=0,
	xmax=1050,
	ymin=0,
	ymax=650,
	ticklabel style={font=\small},
	x label style={at={(axis description cs:0.5,-0.1)},anchor=north},
	y label style={at={(axis description cs:0.20,0.85)},rotate=-90,anchor=south, fill=white},
	axis line style={-{Latex[length=1.5mm,width=1.5mm]}}
]
\coordinate (spypoint) at (axis cs:88,50);
\coordinate (magnifyglass) at (axis cs:800,400);
\draw [thick, dotted, draw=gray] (axis cs: 0,580) -- (axis cs: 1050,580) node[pos=0.9, below] {$\infty$};
\end{axis}
\begin{scope}[spy using outlines={circle, magnification=4, connect spies}]
\node[inner sep=0pt] at (2.17,1.62) {\includegraphics[width=0.37\textwidth]{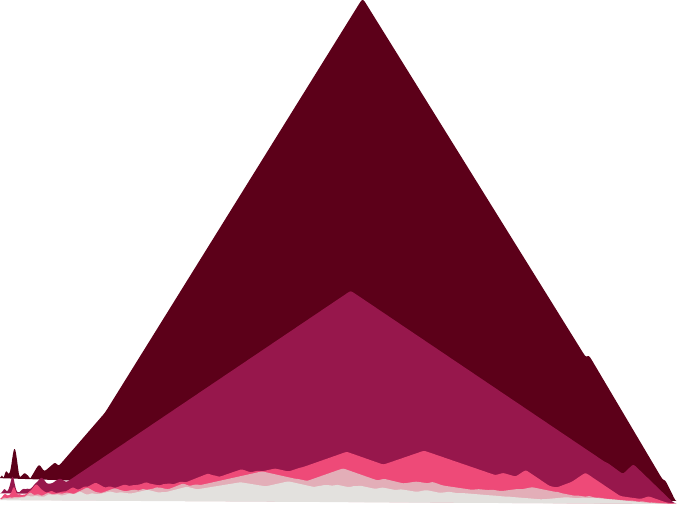}};
\spy [lightgray, size=2cm] on (spypoint) in node[fill=white] at (magnifyglass);
\end{scope} 
\end{tikzpicture}
\begin{tikzpicture}
\begin{axis}[
	height=2in,
	width= 2.4in,
	axis x line = bottom,
	axis y line = left,
	minor x tick num=10,
	xlabel={\textbf{b}) Persistence landscape \texttt{cifar100}.},
	xmin=0,
	xmax=1050,
	ylabel={$\cdot 10^{3}$},
	ymin=0,
	ymax=5,
	ticklabel style={font=\small, fill=white},
	x label style={at={(axis description cs:0.5,-0.1)},anchor=north},
	y label style={at={(axis description cs:0.1,0.9)},rotate=-90,anchor=south},
	axis line style={-{Latex[length=1.5mm,width=1.5mm]}}
]
\draw [thick, dotted, draw=gray] (axis cs: 0,4.45) -- (axis cs: 1050,4.45) node[pos=0.9, below] {$\infty$};
\end{axis}
\node[inner sep=0pt] (h) at (2.17,1.6) {\includegraphics[width=0.37\textwidth]{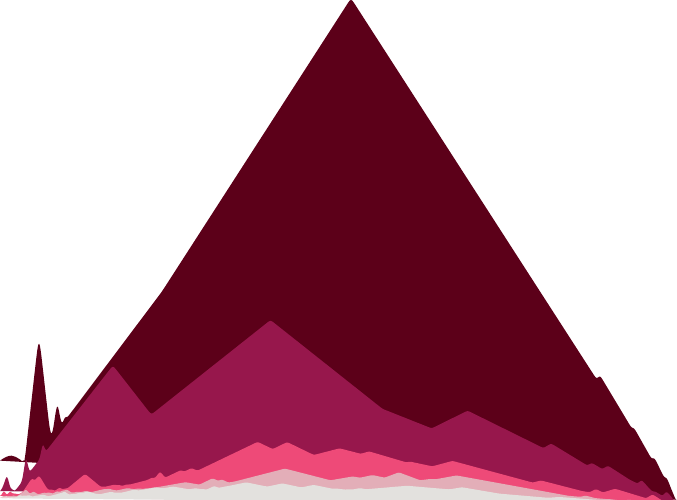}};
\node[inner sep=0pt] (h) at (4.5,2.5) {{\color{red5}$\bullet$} {$H_0$}};
\node[inner sep=0pt] (h) at (4.5,2.1) {{\color{red4}$\bullet$} {$H_1$}};
\node[inner sep=0pt] (h) at (4.5,1.7) {{\color{red3}$\bullet$} {$H_2$}};
\node[inner sep=0pt] (h) at (4.5,1.3) {{\color{red2}$\bullet$} {$H_3$}};
\node[inner sep=0pt] (h) at (4.5,0.9) {{\color{red1}$\bullet$} {$H_4$}};
\end{tikzpicture}
\begin{tikzpicture}
	\begin{axis}[
	table/col sep=comma,
	xlabel={\textbf{c}) Epochs for \texttt{cifar10}.},
	height=2in,
	width= 2.4in,
   	ticklabel style={font=\small, fill=white},
	xticklabel style={rotate=90, anchor=near xticklabel},
	x label style={at={(axis description cs:0.5,-0.15)},anchor=north},
	xtick={0,0.1,0.2,0.4,0.6,0.8,1},
    xticklabels={dummy,0,50,100,150,200,250},
	]
	\end{axis}
	\node[inner sep=0pt] at (2.26,1.87) {\includegraphics[width=0.32\textwidth, height=3.23cm]{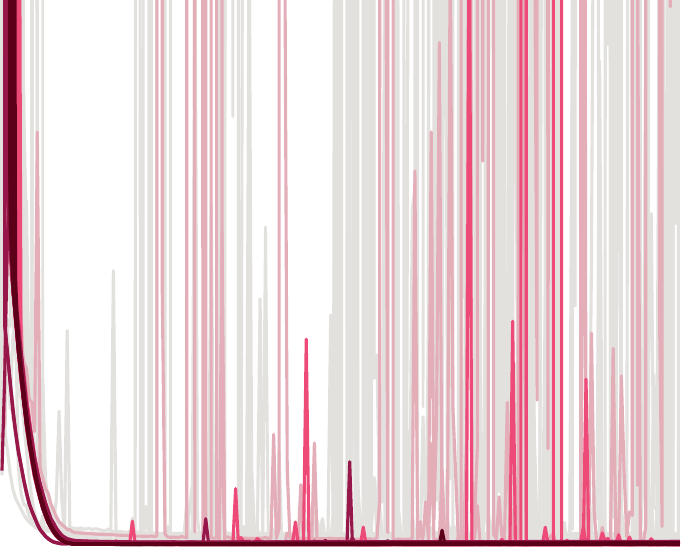}};
\end{tikzpicture}
\begin{tikzpicture}
	\begin{axis}[
	table/col sep=comma,
	xlabel={\textbf{d}) Epochs for \texttt{cifar100}.},
	height=2in,
	width= 2.4in,
	ticklabel style={font=\small, fill=white},
	xticklabel style={rotate=90, anchor=near xticklabel},
	x label style={at={(axis description cs:0.5,-0.15)},anchor=north},
	xtick={0,0.1,0.2,0.4,0.6,0.8,1},
    xticklabels={dummy,0,50,100,150,200,250},
	]
	\end{axis}
	\node[inner sep=0pt] at (2.26,1.87) {\includegraphics[width=0.32\textwidth, height=3.23cm]{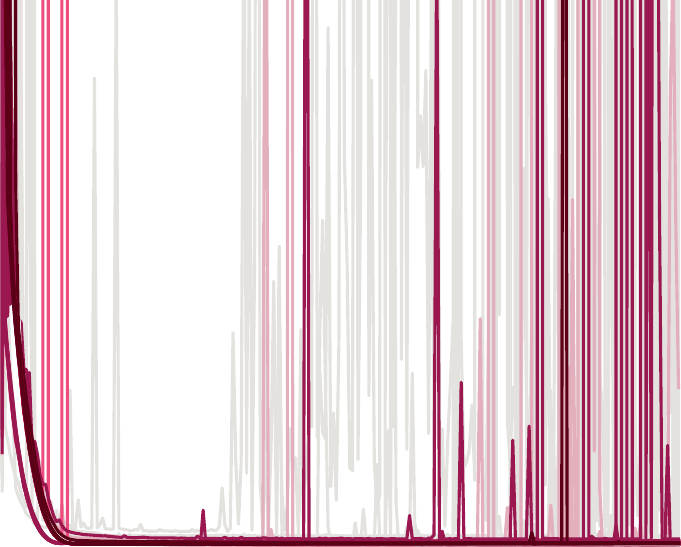}};
\end{tikzpicture}
\caption{a) Persistence landscape for \texttt{cifar10} and b) for the \texttt{cifar100} dataset \cite{krizhevsky2009learning}. Persistence landscapes are computed up to a resolution of $10^3$. c) and d) show the \texttt{MSE} loss function on the validation data using a $7:3$ split on training to test data.}
\label{figure2}
\end{figure}

We train an auto-encoder using as input heavily noisy images and map them to their noiseless original. We perturb each input vector $\textbf{x}^{(0)}$ with Gaussian noise $\epsilon \sim \mathcal{N}(0,\sigma^2 \mathbf{I})$, such that the input is weighted $0.5 \cdot \textbf{x}^{(0)} + 0.5 \cdot \epsilon$. The layer size is increased for each experiment by $2$, i.e. $2,4,6, \cdots, 784$. $392$ neural networks have been trained around a hundred times each. Each line of Fig. \ref{figure2} c), d) represents one neural architecture as averaged loss function.

\subsection{Persistence Landscapes Hyperparameters}
We use the Delaunay-complex for the filtration, according to \cite{lume}. The maximum $\alpha$-square is set to $e^{12}$. The maximal expansion dimension of the simplicial complex is set to $10$. The maximal edge length is set to one. The persistent landscapes are smoothened by a Gaussian filter $G(x) = 1 / \sqrt{2 \pi \sigma^2} \cdot e^{- x^2 / 2 \sigma^2}$ with $\sigma = 2$ for visualization purposes. We implement the persistence diagrams and persistence landscapes using the \texttt{GUDHI v.3.0.0} library -- Geometry Understanding in Higher Dimensions -- with its \texttt{Python}-bindings \cite{gudhi:urm}. Persistent homology is computed on \texttt{Intel Core i}$7$ $9700$\texttt{K} processors.

\subsection{Neural Network Hyperparameters}
We use $\text{L}(\textbf{x}^{(0)}, \textbf{y}) = 1/n \sum_{i=1}^{n} (\textbf{x}^{(0)}_i-\textbf{y}_i)^2$ as loss function. The sigmoid activation function $\sigma(\mathbf{x}^{(l)}) = 1 / (1 + e^{-\mathbf{x}^{(l)}})$ is applied throughout the network to yield the structure of a smooth manifold. Dense neural networks are used with bias term. The data sets are flattened from $(32,32,3)$ into $(3072)$ for the purpose of fitting into a dense layer. The first layer is applied to the flattened data set, which is also adapted by back propagation, i.e. $\varphi^{(0)}(x^{(0)}) =$ $\sigma(\sum_{i=1}^{n-k} \textbf{W}_{ji}^{(1)} \textbf{x}_{i}^{(0)} + \textbf{b}_{i}^{(1)})$. It transfers the data to the desired embedding dimension. The very first and last dense layers are not invertible. Their purpose is a learned projection into the desired subspace of dimension $(n-k)$ for some $k$ and afterwards projecting into the original space in an encoding fashion. The subsequent layers are then implemented as residual invertible layers following Dinh et al. \cite{DinhSB17}.\footnote{Invertible architectures guarantee the same differentiable structure during learning. Due to the construction of trivially invertible neural networks the embedding dimension is doubled, see \cite{DinhSB17}.} A total of $5$ hidden layers are used, all with the same embedding dimension. The batch size is set to $128$, which should help to better recognize good parameterization. Higher batch size causes explosions of the gradient, which we visualize in Fig. \ref{figure2} c), d). For optimization we use \texttt{ADAM} \cite{KingmaB14} and a learning rate of $10^{-4}$. The data set is randomly shuffled. We implement in \texttt{Python v.3.7}. We use \texttt{Tensorflow v.2.2.0} as backend \cite{tensorflow2015-whitepaper} with \texttt{Keras v.2.4.0} \cite{chollet2015keras} as wrapper. The training is conducted on multiple \texttt{NVIDIA Quadro P4000} GPUs.

\subsection{Invertible Neural Network Architecture}
The input vector of any other layer $l$ is given by $\varphi^{(l)} := \textbf{x}^{(l)} = (\textbf{u}^{(l)}_{1},\textbf{u}^{(l)}_{2})$, splitting the $l$-th layer into $\textbf{u}^{(l)}_1$ and $\textbf{u}^{(l)}_2$ \cite{DinhSB17}:
\begin{align}
  \textbf{u}^{(l)}_{1} &= \left(\textbf{v}^{(l)}_1 - \psi_2\left(\textbf{u}^{(l)}_2\right)\right) \cdot \exp\left(-\xi_2\left(\textbf{u}^{(l)}_2\right)\right), \\
  \textbf{u}^{(l)}_{2} &= \left(\textbf{v}^{(l)}_2 - \psi_1\left(\textbf{v}^{(l)}_1\right)\right) \cdot \exp\left(-\xi_1\left(\textbf{v}^{(l)}_1\right)\right).
\end{align}
We apply batch normalization after the activation function. The partition of the vector is still an open research question \cite{DinhSB17}. We divide the flattened vector precisely into two halfs, since the samples are square matrices:
\begin{align}
\mathbf{x}^{(l)} = (\mathbf{u}^{(l)}_1, 0, \cdots, 0) + (0, \cdots, 0, \mathbf{u}^{(l)}_2).
\end{align}
The weights are initialized uniformly distributed and the bias term is with zeros. \texttt{Leaky ReLU} is used as activation function. Batch normalization can be written for a batch $k$ at the $l$-th layer as
\begin{align*}
  \texttt{BN}(\mathbf{x}^{(l)}_{(k)})_{\gamma^{(l)}_{(k)},\beta^{(l)}_{(k)}} = \frac{\gamma^{(l)}_{(k)}}{\sqrt{\text{Var}[\mathbf{x}^{(l)}_{(k)}] + \epsilon^{(l)}_{(k)}}} \mathbf{x}^{(l)}_{(k)} + \left( \beta^{(l)}_{(k)} - \frac{\gamma^{(l)}_{(k)} \text{E}[\mathbf{x}^{(l)}_{(k)}]}{\sqrt{\text{Var}[\mathbf{x}^{(l)}_{(k)}]+\epsilon^{(l)}_{(k)}}} \right),
\end{align*}
where $\gamma^{(l)}_{(k)},\beta^{(l)}_{(k)}$ and $\epsilon^{(l)}_{(k)}$ are parameters updated during back propagation. The two input vectors $\textbf{v}^{(l)}_1$ and $\textbf{v}^{(l)}_2$ of the layer ensure that the output is trivially invertible. The multiplication is to be understood element-wise:
\begin{align}
  \textbf{v}^{(l)}_1 = \textbf{u}^{(l)}_{1} \cdot \exp\left(\xi_2\left(\textbf{u}^{(l)}_{2}\right) + \psi_2\left(\textbf{u}^{(l)}_{2}\right)\right), \\
  \textbf{v}^{(l)}_2 = \textbf{u}^{(l)}_{2} \cdot \exp\left(\xi_1\left(\textbf{v}^{(l)}_{1}\right) + \psi_1\left(\textbf{v}^{(l)}_{1}\right)\right).
\end{align} 
The changes of coordinate charts are Lie group actions from $\text{GL}(n,\mathbb{R})$ on the representation of the first layer. The functions $\xi_1,\xi_2$ and $\psi_1,\psi_2$ are implemented as neural subnetworks themselves and also parameterized during learning. These networks do not have to be invertible in general. Each function is a $\mathcal{C}^1$-residual network with a total of $3$ dense layers \cite{HauserGJR19}. \texttt{Leaky ReLU} has been selected for each layer such that $\texttt{LReLU}(\mathbf{x}^{(l)}_i) = \mathbf{x}^{(l)}_{i}$ if $\mathbf{x}^{(l)}_{i} \geq 0$ and $-5.5 \cdot \mathbf{x}^{(l)}_{i}$ otherwise.

\subsection{Experimental Results}
\begin{table*}[t!]
\centering
\caption{Counts of the representatives per homology group from $\text{Pl}(X)$.}
\label{figure3}
\begin{tabular}{@{}rcccccccccccccc@{}}
\toprule \multirow{2}{*}{\backslashbox{Data \strut}{\strut Features}} & \phantom{abc} & \multicolumn{5}{c}{Homology groups} &&& \multicolumn{6}{c}{$\approx$ embedding dimension}\\
\cmidrule{3-7} \cmidrule{10-15} & & $H_0$ & $H_1$ & $H_2$ & $H_3$ & $H_4$ &&& $p$ & $q \vert {H_1}$ & $q \vert {H_2}$ & $q \vert {H_3}$ & $q \vert {H_4}$ & $\dim U$\\
\midrule
\texttt{cifar10} && 12 & 16 & 40 & 59 & 50 &&& $12$ & $16$ & $9 \scriptstyle{\pm 4}$ & $8 \scriptstyle{\pm 3}$ & $7 \scriptstyle{\pm 15}$ & $92 \scriptstyle{\pm 44}$\\
\texttt{cifar100} && 13 & 18 & 34 & 46 & 48 &&& $13$ & $18$ & $9 \scriptstyle{\pm 2}$ & $8 \scriptstyle{\pm 10}$ & $7 \scriptstyle{\pm 13}$ & $97 \scriptstyle{\pm 50}$\\
\midrule
\end{tabular}
\label{stats}
\end{table*}
We want to represent all invariants of the persistence landscape. This yields the embedding dimensions $\dim U_1 = 92 \scriptstyle{\pm 44}$ for \texttt{cifar10} and $\dim U_2 = 97 \scriptstyle{\pm 50}$ for \texttt{cifar100} (see Tab. \ref{figure3}). According to our assumption, a suitable embedding in an Euclidean space of dimension $2 \cdot \dim U_1 \in [184,272] \subset \mathbb{N}$ and $2 \cdot \dim U_2 \in [194,294] \subset \mathbb{N}$ should be chosen. We interpret, by the similarity of the persistence landscapes (see Fig. \ref{figure1}) and the likewise extremely similar numbers of representatives of the persistent homology groups (see Tab. \ref{figure3}), that the larger data set does differ slightly from a topological viewpoint, i.e. is located in a similar topological space as its smaller counterpart. In Fig. \ref{figure2} c) d) we colored the neural networks according to the chosen embedding, such that the lines show the training of an architecture with the following bottleneck dimensions:
\begin{align}
\texttt{cifar10:} \; {\color{red1}\bullet} \in [2, 148], {\color{red2}\bullet} \in [150, 198], {\color{red3}\bullet} \in [200, 270] \; \text{and} \; {\color{red4}\bullet} \in [272, 784],\\
\texttt{cifar100:} \; {\color{red1}\bullet} \in [2, 148], {\color{red2}\bullet} \in [150, 198], {\color{red3}\bullet} \in [200, 292] \; \text{and} \; {\color{red4}\bullet} \in [294, 784].
\end{align}
All models below our dimensional tresholds $\dim U_i$ -- colored ${\color{red4}\bullet}$ -- show drastic explosions of the gradient (see Fig. \ref{figure2} c), d)). The models above the treshold remain stable. A neural network still shows an explosive gradient in Fig. \ref{figure2} d). It is close to our threshold, identified as $270$ and $292$ dimensional embedding for \texttt{cifar10} and \texttt{cifar100}, respectively. The closer we come to the treshold, the sparser are the fluctuations of loss.

\section{Conclusion}
Based on the theory of Lie groups and persistent homology, a method for neural networks has been developed to parameterize them using the assumption that the data lies on or near by some connected commutative Lie group. This forms an approximate solution for a special case of learning problems. Applying Künneth's theorem, the homology groups of the topological factor spaces could be connected with the ones of their product space. Using persistence landscapes, the elements originating from some homology groups on the filtration were estimated. With numerical experiments we predicted near ideal embedding dimensions and could confirm that a neural embedding above the treshold delivers a loss function on the validation data set with small fluctuations and best reconstruction results (see Fig. \ref{figure2}). We pose following open research questions:
\begin{itemize}
    \item The neural layers do not have the explicit structure of an abelian Lie group. How can spherical CNNs \cite{CohenGKW18} be used to always represent a product space $\mathbb{T}^q \times \mathbb{R}^p $ from layer to layer in order to operate on the proposed Lie group?
    \item This approach can be applied to any kind of manifold, as far as its minimal representation is known. Using decompositions, one may generalize this result. What decomposition allows to neglect the connectedness? For example, the persistence diagrams could be used to identify connected components. These in return could be represented as independent connected manifolds.
    \item Questions come up considering approximations using points from a knotted manifold. How would our method perform in such a case? Can this theory be extended to estimates of embedding dimensions for knotted manifolds?
\end{itemize}

\subsubsection*{Acknowledgements} We thank Christian Holtzhausen, David Haller and Noah Becker for proofreading and anonymous reviewers for their constructive criticism and corrections. This work was partially supported by Siemens Energy AG.

\subsubsection*{Code \& Data} The implementation, the data sets and experimental results can be found at: \href{https://codeberg.org/Jiren/NTOPL}{https://codeberg.org/Jiren/NTOPL}.
\bibliographystyle{splncs04}
\bibliography{biblio}
\end{document}